\documentclass[sigconf]{acmart}
\AtBeginDocument{%
  }

\copyrightyear{2025}
\acmYear{2025}
\setcopyright{cc}
\setcctype{by}
\acmConference[CIKM '25]{Proceedings of the 34th ACM International Conference on Information and Knowledge Management}{November 10--14, 2025}{Seoul, Republic of Korea}
\acmBooktitle{Proceedings of the 34th ACM International Conference on Information and Knowledge Management (CIKM '25), November 10--14, 2025, Seoul, Republic of Korea}\acmDOI{10.1145/3746252.3761650}
\acmISBN{979-8-4007-2040-6/2025/11}

\usepackage{amsmath}
\usepackage{graphicx}
\usepackage{subcaption}
\usepackage{tabularx}
\usepackage{dsfont}
\usepackage{multirow}
\begin{document}

\title{Revisiting Pre-processing Group Fairness: A Modular Benchmarking Framework}

\author{Brodie Oldfield}
\email{brodie.oldfield@data61.csiro.au}
\orcid{0009-0000-4500-5006}
\affiliation{%
	\institution{CSIRO}
	\city{Sydney}
	\country{Australia}
}

\author{Ziqi Xu}
\authornote{Corresponding author.}
\email{ziqi.xu@rmit.edu.au}
\orcid{0000-0003-1748-5801}
\affiliation{%
	\institution{RMIT University}
	\city{Melbourne}
	\country{Australia}}
	
\author{Sevvandi Kandanaarachchi}
\email{sevvandi.kandanaarachchi@data61.csiro.au}
\orcid{0000-0002-0337-0395}
\affiliation{%
	\institution{CSIRO}
	\city{Melbourne}
	\country{Australia}
}


\begin{abstract}
	As machine learning systems become increasingly integrated into high-stakes decision-making processes, ensuring fairness in algorithmic outcomes has become a critical concern. Methods to mitigate bias typically fall into three categories: pre-processing, in-processing, and post-processing. While significant attention has been devoted to the latter two, pre-processing methods, which operate at the data level and offer advantages such as model-agnosticism and improved privacy compliance, have received comparatively less focus and lack standardised evaluation tools. In this work, we introduce {FairPrep}, an extensible and modular benchmarking framework designed to evaluate fairness-aware pre-processing techniques on tabular datasets. Built on the AIF360 platform, FairPrep allows seamless integration of datasets, fairness interventions, and predictive models. It features a batch-processing interface that enables efficient experimentation and automatic reporting of fairness and utility metrics. By offering standardised pipelines and supporting reproducible evaluations, FairPrep fills a critical gap in the fairness benchmarking landscape and provides a practical foundation for advancing data-level fairness research. The source code is available at \url{https://github.com/broldfield/FairPrep}.
\end{abstract}

\ccsdesc[500]{Computing methodologies~Machine learning}
\ccsdesc[500]{General and reference~Evaluation}

\keywords{Machine Learning, Fairness, Pre-processing}
%

\maketitle

\section{Introduction}
Machine learning techniques are becoming increasingly prevalent, shaping how we interact with data and influencing decision-making across diverse application areas. As these techniques are deployed in scenarios where decisions can significantly impact individual livelihoods, such as employment, finance, and healthcare, the question of fairness in automated decisions becomes essential. In particular, we must ask whether algorithms systematically treat individuals differently based on sensitive attributes such as race or gender~\cite{LuoHYHH0024}.


Fairness-aware machine learning methods are commonly divided into three categories: pre-processing, in-processing, and post-processing. Pre-processing techniques modify the training data to reduce bias, such as through re-weighting, resampling, or transforming features to obscure sensitive attributes~\cite{KamiranC11,CalmonWVRV17,fairrepresentations,disparateImpactPaper}. In-processing methods embed fairness constraints into the training process, often via regularisation or adversarial objectives that penalise group disparities~\cite{KamishimaAAS12,KearnsNRW18,AgarwalBD0W18,AgarwalDW19,XuLCLLW23,zhao2025fairdrl}. Post-processing techniques adjust model outputs after training, for example by shifting decision thresholds to equalise outcomes across groups~\cite{PleissRWKW17,HardtPNS16,KamiranKZ12,GeyikAK19}. These approaches differ in terms of model access, flexibility, and applicability, with pre-processing being especially appealing due to its simplicity and model-agnostic design.

Despite these established categories, prior research has mainly focused on in-processing and post-processing techniques, while pre-processing methods have received less attention. This discrepancy is largely due to the difficulty of standardising evaluation for pre-processing approaches, as many fairness benchmarks lack support for flexible and transparent data manipulation. Tools such as AIF360~\cite{aif360-oct-2018}, UST~\cite{XuXLCLLW22}, and FairLearn~\cite{bird2020fairlearn} focus on model-level adjustments. By contrast, Fair-IRT~\cite{XuKON25} builds on broader research leveraging item response theory (IRT) for algorithm evaluation~\cite{OldfieldKXM25, XuMRC0X25}. Fair-IRT continues this trajectory by explicitly integrating fairness considerations, bridging both model- and data-level evaluation. In addition, inconsistencies in pre-processing, such as outlier removal strategies or methods for splitting training and testing sets, can cause substantial differences in reported performance. Even small changes in data preparation may yield significant variations in outcomes, raising concerns about the reliability and reproducibility of fairness evaluations~\cite{han2024ffbfairfairnessbenchmark}.

To address the lack of standardised evaluation for pre-processing fairness methods, we introduce {FairPrep}, an extensible and modular framework for benchmarking fairness-aware data transformation techniques on tabular datasets. Built on top of the AIF360 foundation, FairPrep is designed with usability and research flexibility in mind. It enables users to seamlessly integrate new datasets, fairness techniques, and predictive models through a modular API. To facilitate large-scale and reproducible experiments, the framework includes a batch-processing interface that allows users to define experimental configurations and automatically compute both fairness and utility metrics. FairPrep contributes to the fairness research community by: (1) standardising pre-processing evaluation across datasets and metrics, (2) supporting easy extension and integration with new methods, and (3) streamlining the benchmarking pipeline for reproducible research. To the best of our knowledge, this is the first benchmarking study focused specifically on group fairness through pre-processing techniques.

\section{The proposed FairPrep}
This section presents FairPrep, a modular and extensible benchmarking framework for fairness-aware pre-processing. We introduce its two-stage architecture, describe the fairness metrics, outline the five benchmark datasets, and detail the four integrated pre-processing techniques. 

\subsection{System Architecture}

FairPrep operates in two stages: a pre-processing stage and a benchmarking stage. These are executed independently via command-line interfaces, allowing modular and flexible usage. In the pre-processing stage, the user specifies a dataset, a sensitive attribute, and a fairness pre-processing method. The technique is applied to transform the dataset and reduce bias with respect to the sensitive attribute. Fairness metrics are computed for both the original and processed datasets to assess the effect of the transformation. Both versions are cached for reuse in benchmarking. In the benchmarking stage, a predictive model is trained on both datasets using a holdout validation scheme (e.g., 70\% training, 15\% validation, 15\% testing). Performance and fairness metrics are computed from the predictions to evaluate trade-offs. To support fine-grained analysis, thresholds are swept from 0.01 to 0.99 in increments of 0.01. At each threshold, metrics are recorded, and the optimal threshold balancing fairness and accuracy is identified and visualised.

To enable scalable experiments, FairPrep supports YAML-based batch execution. Users may define fixed jobs or provide lists of inputs to automatically generate all valid combinations. This supports reproducible and extensible benchmarking across datasets, models, and fairness methods.

\subsection{Group Fairness Metrics}
FairPrep adopts a set of comprehensive metrics to assess data-level fairness before training and model-level fairness and utility after training. These metrics are grouped by stage: pre-processing or benchmarking.

\subsubsection{Metrics for Pre-processing Evaluation}

In the pre-processing stage, FairPrep evaluates the effectiveness of fairness interventions using several group-level and distributional metrics. These include: (1) \textit{base rate}, which measures the overall proportion of positive labels, and its conditional variant $\Pr(Y = 1 \mid S = s)$; (2) \textit{consistency}, which assesses individual-level fairness by measuring the agreement between a sample’s label and those of its neighbors in the feature space; (3) \textit{disparate impact}, and (4) \textit{statistical parity}, which capture group-level fairness by quantifying outcome ratios and differences across sensitive groups; (5) \textit{number of positives} and (6) \textit{number of negatives}, which are computed as the total counts of positive and negative labels, respectively, providing a basic view of label distributions; and (7) \textit{empirical difference}, which quantifies the disparity of a selected metric between privileged and unprivileged groups. Together, these metrics offer a comprehensive assessment of data-level bias prior to model training.

\subsubsection{Metrics for Benchmarking Evaluation}

In the benchmarking stage, FairPrep evaluates the predictive models trained on both original and pre-processed datasets using a combination of performance and fairness metrics. Utility is assessed using (8)~\textit{balanced accuracy}, defined as the average of the true positive rate (TPR) and true negative rate (TNR), which is particularly useful in imbalanced classification tasks. Fairness is evaluated using several group-based metrics, including (9)~\textit{equal opportunity}, which measures the difference in TPR between underprivileged and privileged groups, and (10)~\textit{equal odds}, which averages disparities in both TPR and false positive rate (FPR) across groups. For consistency with the pre-processing evaluation, the stage also computes \textit{disparate impact} and \textit{statistical parity difference}. Additionally, (11)~\textit{the theil index}, a generalised entropy-based metric with parameter $\alpha = 1$~\cite{speicher2018unified}, is included to capture both individual- and group-level fairness violations. Together, these benchmarking metrics provide a comprehensive evaluation of how fairness-aware pre-processing techniques affect downstream model behaviour.

\subsection{Datasets}

We evaluate FairPrep on 5 commonly used fairness-related datasets, briefly described below: (1) the \textit{Adult Census} dataset~\cite{adult_2} contains demographic and financial information of individuals from the U.S. Census Bureau, with the target variable indicating whether an individual earns over \$50K annually; the sensitive attribute is typically gender or race; (2) the \textit{Bank Marketing} dataset~\cite{bank_marketing_222} includes client information and marketing interaction data collected by a Portuguese banking institution; the target variable denotes whether a client subscribes to a term deposit, with age or marital status used as the sensitive attribute; (3) the \textit{ProPublica COMPAS} dataset~\cite{propublica_compas} comprises criminal history and demographic attributes of defendants; the prediction task focuses on recidivism within two years, and race is commonly used as the sensitive attribute; (4) the \textit{German Credit} dataset~\cite{statlog_(german_credit_data)_144} contains features from credit applicants at a German bank, with the target being a binary indicator of credit risk (good or bad); gender or age is often used as the sensitive attribute; and (5) the \textit{MEPS Panel 21} dataset~\cite{meps21} is drawn from the Medical Expenditure Panel Survey, which includes data on demographics, healthcare usage, costs, and quality of care collected from individuals, providers, and employers in the United States. In this dataset, the target variable is \textit{utilization}, a score reflecting the appropriateness and frequency of care received. We binarize this score such that values less than 10 are labelled as 0 (low utilisation), and values greater than or equal to 10 are labelled as 1 (high utilisation), with race used as the sensitive attribute.

\begin{table*}[t]
	\centering
	\caption{Results of fairness metrics in the pre-processing stage for 3 datasets and 4 methods.}
	\label{tab:pp_metrics_summary}
	\setlength\tabcolsep{4pt}
	\small{\begin{tabular}{|l|ccccc|ccccc|ccccc|}
			\toprule
			& \multicolumn{5}{c|}{Adult Census} & \multicolumn{5}{c|}{ProPublica COMPAS} & \multicolumn{5}{c|}{German Credit} \\ \midrule
			Metrics & Orig. & RW & LFR & DIR & OPP & Orig. & RW & LFR & DIR & OPP & Orig. & RW & LFR & DIR & OPP \\ \midrule
			(1) base rate            & 0.239 & 0.239 & 0.095 & 0.239 & 0.252 & 0.530 & 0.530 & 0.031 & 0.530 & 0.529 & 0.700 & 0.700 & 1 & 0.700 & 0.694 \\
			(2) consistency          & 0.719 & 0.719 & 1 & 0.719 & 0.700 & 0.595 & 0.595 & 1 & 0.595 & 0.610 & 0.661 & 0.661 & 1 & 0.661 & 0.644 \\
			(3) disparate impact     & 0.360 & 1.000 & 0.601   & 0.360 & 0.816 & 0.788 & 1.000 & 0.786 & 0.788 & 0.906 & 0.897 & 1.000 & 1 & 0.897 & 0.927 \\
			(4) statistical parity   & -0.195 & 0.000 & -0.043 & -0.195 & -0.050 & -0.135 & 0.000 & -0.008 & -0.135 & -0.054 & -0.075 & 0.000 & 0 & -0.075 & -0.052 \\
			(5) Num. positives  & 11687 & 11687 & 4664 & 11687 & 12330 & 2795 & 2795 & 161 & 2795 & 2793 & 700 & 700 & 1000 & 700 & 694 \\
			(6) Num. negatives  & 37155 & 37155 & 44178 & 37155 & 36512 & 2483 & 2483 & 5117  & 2483 & 2485 & 300 & 300 & 0 & 300 & 306 \\
			(7) empirical difference & 1.022 & 0.000 & 0.509 & 1.022 & 0.205 & 0.317 & 0.000 & 0.249 & 0.317 & 0.118 & 0.239 & 0.001 & 0.798 & 0.239 & 0.166 \\ \bottomrule
	\end{tabular}}
\end{table*}

\subsection{Fairness Pre-processing Methods}

The proposed FairPrep involve four widely studied fairness pre-processing methods: Reweighing, Learned Fair Representations, Disparate Impact Remover, and Optimised Pre-processing. We consider a dataset $D = (S, X, Y)$, with protected attribute $S$, features $X$, and binary label $Y$ in the following discussion.  

\subsubsection{Reweighing (RW)~\cite{KamiranC11}} RW assigns weights to samples based on the observed and expected probabilities of each (group, label) pair, assuming independence between group membership and outcome. For a dataset $D = (S, X, Y)$, the weight for a given pair $(s, y)$ is computed as:
\[
W(s, y) = \frac{Pr(S = s) \times Pr(Y = y)}{Pr(S = s \wedge Y = y)}.
\]
This adjusts the training distribution to reduce the disparity between groups. For example, if the observed success rate of a group is lower than expected, its positive instances will be up-weighted accordingly. These weights are passed to the classifier and used to reweight the loss during training.

\subsubsection{Learned Fair Representations (LFR)~\cite{fairrepresentations}} LFR learns a latent representation of the input data that retains task-relevant information while obfuscating sensitive attribute information~\cite{fairrepresentations}. Given a dataset $D = (S, X, Y)$, the goal is to find a prototype set $Z$ such that: (1) the mapping from $X$ to $Z$ satisfies statistical parity; (2) $Z$ retains as much information as possible unrelated to $S$; and (3) the composed mapping from $X$ to $Z$ and then to $Y$ approximates the original classifier $f: X \rightarrow Y$.

\subsubsection{Disparate Impact Remover (DIR)~\cite{disparateImpactPaper}} DIR edits feature values to improve group fairness while preserving the rank-ordering of features within groups~\cite{disparateImpactPaper}. The concept of disparate impact has legal origins and is defined as:

\begin{definition}
	Given a dataset $D = (S, X, Y)$, with protected attribute $S$, features $X$, and binary label $Y$, disparate impact exists if
	\[
	\frac{Pr(Y = 1 \mid S = 0)}{Pr(Y = 1 \mid S = 1)} \leq \tau = 0.8,
	\]
	for positive outcomes $Y = 1$ and majority group $S = 1$.
\end{definition}

DIR produces a repaired version of $X$, denoted $\tilde{X}$, such that the marginal distributions maintain the same cumulative probabilities. Formally, if $F_s: X_s \rightarrow [0,1]$ is the cumulative distribution function for group $S = s$, then the repaired features $\tilde{x}$ satisfy $F_s(x) = F_s(\tilde{x})$.

\subsubsection{Optimised Pre-processing (OPP)~\cite{CalmonWVRV17}} OPP learns a probabilistic transformation of both features and labels to minimise discrimination while preserving individual utility and limiting distortion~\cite{CalmonWVRV17}. Given a dataset $D = (S, X, Y)$, the transformed dataset $\hat{D} = (S, \hat{X}, \hat{Y})$ is constructed such that the joint distribution $(\hat{X}, \hat{Y})$ is close to the original $(X, Y)$, while ensuring that $Pr(\hat{Y} \mid S)$ is close to a target distribution and that the changes to individual data points are small.

\begin{figure*}[t]
	\centering
	\begin{subfigure}[b]{\linewidth}
		\centerline{\includegraphics[width=\linewidth]{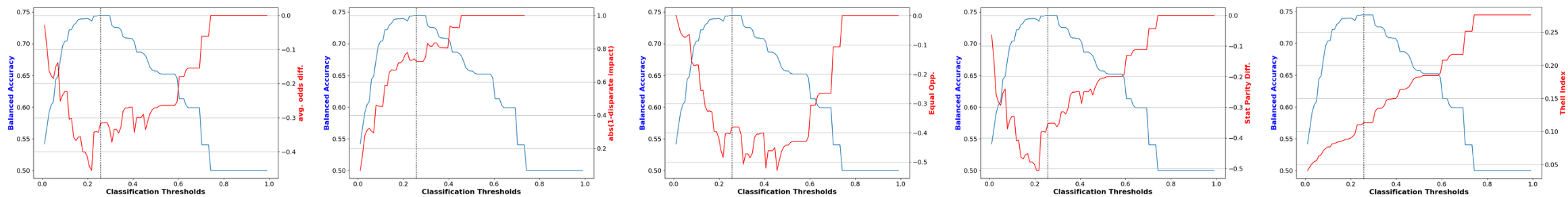}}
		\caption{}
		\label{Adult_RW_Original}
	\end{subfigure}
	\begin{subfigure}[b]{\linewidth}
		\centerline{\includegraphics[width=\linewidth]{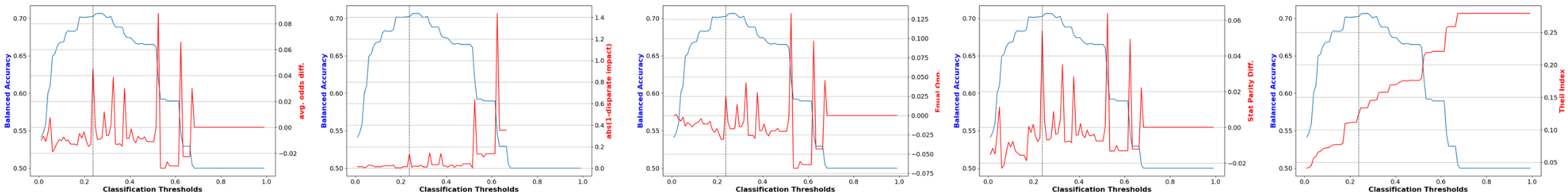}}
		\caption{}
		\label{Adult_RW_Processed}
	\end{subfigure}
	\caption{Balanced Accuracy (left axis, blue) versus fairness metrics (right axis, red) across varying classification thresholds for five different fairness metrics. Each subplot corresponds to a specific metric. 
		(a) Results obtained from the original Adult dataset. 
		(b) Results obtained after applying the Reweighing pre-processing method to the Adult dataset. 
		Logistic Regression is used as the predictive model in both settings.}
	\label{Adult_combined}
\end{figure*}

\section{Experiments}
In this section, we present empirical evaluations of FairPrep to demonstrate its flexibility, modular design, and practical utility in benchmarking fairness-aware pre-processing techniques. Applied to multiple benchmark datasets, the results show how different techniques influence group fairness and confirm FairPrep’s ability to support standardised, reproducible evaluation.

\subsection{Experimental Setup}
To showcase the practicality and extensibility of \textit{FairPrep}, we conduct experiments using five commonly studied tabular datasets and four representative fairness pre-processing techniques. These datasets are processed using our pre-processing interface and evaluated via predictive models. It is important to note that FairPrep is designed as a modular and extensible framework. All components, including datasets, fairness pre-processing techniques, and predictive models, are exposed via standardised interfaces. This allows users to easily plug in their own datasets, implement custom pre-processing methods, or evaluate new classifiers without modifying the core framework. The experiments presented here serve as instantiations of this general pipeline.

Our evaluation follows a two-stage pipeline. In the pre-processing stage, the framework computes pre-processing metrics on both the original and transformed datasets to assess the effect of bias mitigation before model training. These results are summarised in Table~\ref{tab:pp_metrics_summary}. In the benchmarking stage, each model is trained on the original and pre-processed datasets using a holdout validation scheme, and evaluated on both fairness and performance metrics. The resulting metrics are used to generate threshold-sensitive trade-off plots (e.g., Figures~\ref{Adult_RW_Original} and~\ref{Adult_RW_Processed}), which offer a fine-grained view of how fairness and utility evolve across decision thresholds. 

Due to space limitations, we present the results from the pre-processing stage for three datasets and four fairness-aware pre-processing methods. For the benchmarking stage, we report the comparative results of the original and processed versions of the Adult dataset using Logistic Regression. The full experimental results can be reproduced via the link provided in the abstract.

\subsection{Results for Pre-processing Stage}
Table~\ref{tab:pp_metrics_summary} summarises fairness metrics computed on the original and processed datasets across three benchmark datasets and four pre-processing methods. Several consistent trends emerge. (1) RW improves group fairness across all datasets, achieving perfect disparate impact (1.0) and eliminating statistical parity difference (0.0), while leaving label distributions unchanged. This outcome aligns with its design, which adjusts instance weights without modifying features or labels, making it a practical choice when preserving original data semantics is important. (2) LFR introduces substantial changes to the data, often removing all positive labels (Adult, COMPAS) or setting all to positive (German). While this results in ideal fairness metrics, such extreme shifts make the data unrealistic and potentially unusable for downstream tasks. The high empirical differences further indicate that fairness is achieved at the cost of data fidelity. (3) DIR produces minimal change in group-level fairness metrics, with little to no improvement in disparate impact or statistical parity. This suggests that its rank-preserving transformations alone are insufficient for correcting entrenched disparities. (4) OPP strikes a more balanced trade-off by improving fairness metrics while maintaining a reasonable label distribution. Its empirical differences are significantly lower than those in the original data, indicating more stable and effective bias mitigation. 

In summary, RW and OPP offer controlled and practical improvements, whereas LFR, despite its effectiveness on fairness metrics, may compromise data integrity. DIR shows limited standalone impact and may benefit from being used in conjunction with other techniques.

\subsection{Results for Benchmarking Stage}

Figure~\ref{Adult_combined} illustrates how balanced accuracy and various fairness metrics evolve as classification thresholds change, using the Adult dataset and Logistic Regression as the predictive model. Figure~\ref{Adult_RW_Original} presents the results on the original dataset, which exhibits substantial variability and sensitivity to threshold selection. Fairness metrics such as equal opportunity and average odds difference fluctuate considerably, especially in mid-range thresholds. The Theil index increases sharply at more lenient thresholds, indicating rising outcome inequality. These trends suggest that fairness behaviour on the original dataset is highly unstable and threshold-dependent, complicating model deployment in practice.

By contrast, Figure~\ref{Adult_RW_Processed} shows the corresponding results after applying the RW. The processed dataset yields notably better fairness trends despite some fluctuations. Most metrics, including statistical parity difference and disparate impact, remain closer to zero across the threshold range, reflecting improved group-level parity. Although some residual fluctuations persist in average odds difference, the overall behaviour is more robust. Importantly, balanced accuracy remains comparable to the original, indicating that fairness gains are achieved without sacrificing model utility. These findings highlight RW’s effectiveness not only in improving fairness metrics at fixed thresholds but also in enhancing the stability of fairness-utility trade-offs across decision boundaries.

\section{Conclusion}

In this paper, we present FairPrep, a modular and extensible benchmarking framework for fairness-aware pre-processing in tabular data. By supporting a wide range of datasets, transformation techniques, and predictive models, FairPrep provides a unified and reproducible pipeline for evaluating the trade-offs between fairness and utility. Our empirical results demonstrate the distinct impacts of common pre-processing methods on both data characteristics and downstream model behaviour. While FairPrep standardises evaluation and simplifies extensibility, one limitation is its focus on binary classification tasks with group fairness metrics. Future work will extend the framework to accommodate multi-class and regression tasks, incorporate individual fairness measures, and support integration with modern deep learning pipelines.

\begin{acks}
	This work was supported by the research support package from the School of Computing Technologies at RMIT University.
\end{acks}

\section*{GenAI Usage Disclosure}
During this research, generative AI tools are used to enhance the quality and efficiency of specific tasks, all within the permitted scope defined by the ACM policies on the use of GenAI. These tools primarily assist in refining the manuscript's linguistic elements, such as grammar, syntax, and minor stylistic adjustments, to ensure clarity and conciseness. Additionally, during the implementation phase, AI suggests relevant programming libraries and provides guidance on Python syntax, which helps accelerate development while following established coding practices. All uses are strictly limited to supporting functions and do not involve the generation or interpretation of research data, methodology, or conclusions.

\bibliographystyle{ACM-Reference-Format}
\balance
\bibliography{reference.bib}


\begin{thebibliography}{28}


\ifx \showCODEN    \undefined \def \showCODEN     #1{\unskip}     \fi
\ifx \showISBNx    \undefined \def \showISBNx     #1{\unskip}     \fi
\ifx \showISBNxiii \undefined \def \showISBNxiii  #1{\unskip}     \fi
\ifx \showISSN     \undefined \def \showISSN      #1{\unskip}     \fi
\ifx \showLCCN     \undefined \def \showLCCN      #1{\unskip}     \fi
\ifx \shownote     \undefined \def \shownote      #1{#1}          \fi
\ifx \showarticletitle \undefined \def \showarticletitle #1{#1}   \fi
\ifx \showURL      \undefined \def \showURL       {\relax}        \fi
\providecommand\bibfield[2]{#2}
\providecommand\bibinfo[2]{#2}
\providecommand\natexlab[1]{#1}
\providecommand\showeprint[2][]{arXiv:#2}

\bibitem[Agarwal et~al\mbox{.}(2018)]%
        {AgarwalBD0W18}
\bibfield{author}{\bibinfo{person}{Alekh Agarwal}, \bibinfo{person}{Alina
  Beygelzimer}, \bibinfo{person}{Miroslav Dud{\'{\i}}k}, \bibinfo{person}{John
  Langford}, {and} \bibinfo{person}{Hanna~M. Wallach}.}
  \bibinfo{year}{2018}\natexlab{}.
\newblock \showarticletitle{A Reductions Approach to Fair Classification}. In
  \bibinfo{booktitle}{\emph{Proceedings of the 35th International Conference on
  Machine Learning, {ICML} 2018}}, Vol.~\bibinfo{volume}{80}.
  \bibinfo{pages}{60--69}.
\newblock
\urldef\tempurl%
\url{http://proceedings.mlr.press/v80/agarwal18a.html}
\showURL{%
\tempurl}


\bibitem[Agarwal et~al\mbox{.}(2019)]%
        {AgarwalDW19}
\bibfield{author}{\bibinfo{person}{Alekh Agarwal}, \bibinfo{person}{Miroslav
  Dud{\'{\i}}k}, {and} \bibinfo{person}{Zhiwei~Steven Wu}.}
  \bibinfo{year}{2019}\natexlab{}.
\newblock \showarticletitle{Fair Regression: Quantitative Definitions and
  Reduction-Based Algorithms}. In \bibinfo{booktitle}{\emph{Proceedings of the
  36th International Conference on Machine Learning, {ICML} 2019}},
  Vol.~\bibinfo{volume}{97}. \bibinfo{pages}{120--129}.
\newblock
\urldef\tempurl%
\url{http://proceedings.mlr.press/v97/agarwal19d.html}
\showURL{%
\tempurl}


\bibitem[{Agency for Healthcare Research and Quality (AHRQ)}(2016)]%
        {meps21}
\bibfield{author}{\bibinfo{person}{{Agency for Healthcare Research and Quality
  (AHRQ)}}.} \bibinfo{year}{2016}\natexlab{}.
\newblock \bibinfo{title}{Medical Expenditure Panel Survey (MEPS) Panel 21
  Longitudinal Data File}.
\newblock
\urldef\tempurl%
\url{https://meps.ahrq.gov/mepsweb/data_stats/download_data_files.jsp}
\showURL{%
\tempurl}
\newblock
\shownote{U.S. Department of Health \& Human Services}.


\bibitem[Becker and Kohavi(1996)]%
        {adult_2}
\bibfield{author}{\bibinfo{person}{Barry Becker} {and} \bibinfo{person}{Ronny
  Kohavi}.} \bibinfo{year}{1996}\natexlab{}.
\newblock \bibinfo{title}{{Adult}}.
\newblock \bibinfo{howpublished}{UCI Machine Learning Repository}.
\newblock
\newblock
\shownote{{DOI}: https://doi.org/10.24432/C5XW20}.


\bibitem[Bellamy et~al\mbox{.}(2018)]%
        {aif360-oct-2018}
\bibfield{author}{\bibinfo{person}{Rachel K.~E. Bellamy},
  \bibinfo{person}{Kuntal Dey}, \bibinfo{person}{Michael Hind},
  \bibinfo{person}{Samuel~C. Hoffman}, \bibinfo{person}{Stephanie Houde},
  \bibinfo{person}{Kalapriya Kannan}, \bibinfo{person}{Pranay Lohia},
  \bibinfo{person}{Jacquelyn Martino}, \bibinfo{person}{Sameep Mehta},
  \bibinfo{person}{Aleksandra Mojsilovic}, \bibinfo{person}{Seema Nagar},
  \bibinfo{person}{Karthikeyan~Natesan Ramamurthy}, \bibinfo{person}{John
  Richards}, \bibinfo{person}{Diptikalyan Saha}, \bibinfo{person}{Prasanna
  Sattigeri}, \bibinfo{person}{Moninder Singh}, \bibinfo{person}{Kush~R.
  Varshney}, {and} \bibinfo{person}{Yunfeng Zhang}.}
  \bibinfo{year}{2018}\natexlab{}.
\newblock \bibinfo{title}{{AI Fairness} 360: An Extensible Toolkit for
  Detecting, Understanding, and Mitigating Unwanted Algorithmic Bias}.
\newblock
\urldef\tempurl%
\url{https://arxiv.org/abs/1810.01943}
\showURL{%
\tempurl}


\bibitem[Calmon et~al\mbox{.}(2017)]%
        {CalmonWVRV17}
\bibfield{author}{\bibinfo{person}{Fl{\'{a}}vio~P. Calmon},
  \bibinfo{person}{Dennis Wei}, \bibinfo{person}{Bhanukiran Vinzamuri},
  \bibinfo{person}{Karthikeyan~Natesan Ramamurthy}, {and}
  \bibinfo{person}{Kush~R. Varshney}.} \bibinfo{year}{2017}\natexlab{}.
\newblock \showarticletitle{Optimized Pre-Processing for Discrimination
  Prevention}. In \bibinfo{booktitle}{\emph{Advances in Neural Information
  Processing Systems 30: Annual Conference on Neural Information Processing
  Systems 2017}}. \bibinfo{pages}{3992--4001}.
\newblock
\urldef\tempurl%
\url{https://proceedings.neurips.cc/paper/2017/hash/9a49a25d845a483fae4be7e341368e36-Abstract.html}
\showURL{%
\tempurl}


\bibitem[Feldman et~al\mbox{.}(2015)]%
        {disparateImpactPaper}
\bibfield{author}{\bibinfo{person}{Michael Feldman},
  \bibinfo{person}{Sorelle~A. Friedler}, \bibinfo{person}{John Moeller},
  \bibinfo{person}{Carlos Scheidegger}, {and} \bibinfo{person}{Suresh
  Venkatasubramanian}.} \bibinfo{year}{2015}\natexlab{}.
\newblock \showarticletitle{Certifying and Removing Disparate Impact}. In
  \bibinfo{booktitle}{\emph{Proceedings of the 21th {ACM} {SIGKDD}
  International Conference on Knowledge Discovery and Data Mining, {KDD}
  2015}}. \bibinfo{pages}{259--268}.
\newblock
\href{https://doi.org/10.1145/2783258.2783311}{doi:\nolinkurl{10.1145/2783258.2783311}}


\bibitem[Geyik et~al\mbox{.}(2019)]%
        {GeyikAK19}
\bibfield{author}{\bibinfo{person}{Sahin~Cem Geyik}, \bibinfo{person}{Stuart
  Ambler}, {and} \bibinfo{person}{Krishnaram Kenthapadi}.}
  \bibinfo{year}{2019}\natexlab{}.
\newblock \showarticletitle{Fairness-Aware Ranking in Search {\&}
  Recommendation Systems with Application to LinkedIn Talent Search}. In
  \bibinfo{booktitle}{\emph{Proceedings of the 25th {ACM} {SIGKDD}
  International Conference on Knowledge Discovery {\&} Data Mining, {KDD}
  2019}}. \bibinfo{publisher}{{ACM}}, \bibinfo{pages}{2221--2231}.
\newblock
\href{https://doi.org/10.1145/3292500.3330691}{doi:\nolinkurl{10.1145/3292500.3330691}}


\bibitem[Han et~al\mbox{.}(2024)]%
        {han2024ffbfairfairnessbenchmark}
\bibfield{author}{\bibinfo{person}{Xiaotian Han}, \bibinfo{person}{Jianfeng
  Chi}, \bibinfo{person}{Yu Chen}, \bibinfo{person}{Qifan Wang},
  \bibinfo{person}{Han Zhao}, \bibinfo{person}{Na Zou}, {and}
  \bibinfo{person}{Xia Hu}.} \bibinfo{year}{2024}\natexlab{}.
\newblock \showarticletitle{{FFB:} {A} Fair Fairness Benchmark for
  In-Processing Group Fairness Methods}. In \bibinfo{booktitle}{\emph{The
  Twelfth International Conference on Learning Representations, {ICLR} 2024}}.
\newblock
\urldef\tempurl%
\url{https://openreview.net/forum?id=TzAJbTClAz}
\showURL{%
\tempurl}


\bibitem[Hardt et~al\mbox{.}(2016)]%
        {HardtPNS16}
\bibfield{author}{\bibinfo{person}{Moritz Hardt}, \bibinfo{person}{Eric Price},
  {and} \bibinfo{person}{Nati Srebro}.} \bibinfo{year}{2016}\natexlab{}.
\newblock \showarticletitle{Equality of Opportunity in Supervised Learning}. In
  \bibinfo{booktitle}{\emph{Advances in Neural Information Processing Systems
  29: Annual Conference on Neural Information Processing Systems 2016}}.
  \bibinfo{pages}{3315--3323}.
\newblock
\urldef\tempurl%
\url{https://proceedings.neurips.cc/paper/2016/hash/9d2682367c3935defcb1f9e247a97c0d-Abstract.html}
\showURL{%
\tempurl}


\bibitem[Hofmann(1994)]%
        {statlog_(german_credit_data)_144}
\bibfield{author}{\bibinfo{person}{Hans Hofmann}.}
  \bibinfo{year}{1994}\natexlab{}.
\newblock \bibinfo{title}{{Statlog (German Credit Data)}}.
\newblock \bibinfo{howpublished}{UCI Machine Learning Repository}.
\newblock
\newblock
\shownote{{DOI}: https://doi.org/10.24432/C5NC77}.


\bibitem[Kamiran and Calders(2011)]%
        {KamiranC11}
\bibfield{author}{\bibinfo{person}{Faisal Kamiran} {and} \bibinfo{person}{Toon
  Calders}.} \bibinfo{year}{2011}\natexlab{}.
\newblock \showarticletitle{Data preprocessing techniques for classification
  without discrimination}.
\newblock \bibinfo{journal}{\emph{Knowledge and Information Systems}}
  \bibinfo{volume}{33}, \bibinfo{number}{1} (\bibinfo{year}{2011}),
  \bibinfo{pages}{1--33}.
\newblock
\href{https://doi.org/10.1007/S10115-011-0463-8}{doi:\nolinkurl{10.1007/S10115-011-0463-8}}


\bibitem[Kamiran et~al\mbox{.}(2012)]%
        {KamiranKZ12}
\bibfield{author}{\bibinfo{person}{Faisal Kamiran}, \bibinfo{person}{Asim
  Karim}, {and} \bibinfo{person}{Xiangliang Zhang}.}
  \bibinfo{year}{2012}\natexlab{}.
\newblock \showarticletitle{Decision Theory for Discrimination-Aware
  Classification}. In \bibinfo{booktitle}{\emph{12th {IEEE} International
  Conference on Data Mining, {ICDM} 2012}}. \bibinfo{pages}{924--929}.
\newblock
\href{https://doi.org/10.1109/ICDM.2012.45}{doi:\nolinkurl{10.1109/ICDM.2012.45}}


\bibitem[Kamishima et~al\mbox{.}(2012)]%
        {KamishimaAAS12}
\bibfield{author}{\bibinfo{person}{Toshihiro Kamishima},
  \bibinfo{person}{Shotaro Akaho}, \bibinfo{person}{Hideki Asoh}, {and}
  \bibinfo{person}{Jun Sakuma}.} \bibinfo{year}{2012}\natexlab{}.
\newblock \showarticletitle{Fairness-Aware Classifier with Prejudice Remover
  Regularizer}. In \bibinfo{booktitle}{\emph{Machine Learning and Knowledge
  Discovery in Databases - European Conference, {ECML} {PKDD} 2012}},
  Vol.~\bibinfo{volume}{7524}. \bibinfo{pages}{35--50}.
\newblock
\href{https://doi.org/10.1007/978-3-642-33486-3\_3}{doi:\nolinkurl{10.1007/978-3-642-33486-3\_3}}


\bibitem[Kearns et~al\mbox{.}(2018)]%
        {KearnsNRW18}
\bibfield{author}{\bibinfo{person}{Michael~J. Kearns}, \bibinfo{person}{Seth
  Neel}, \bibinfo{person}{Aaron Roth}, {and} \bibinfo{person}{Zhiwei~Steven
  Wu}.} \bibinfo{year}{2018}\natexlab{}.
\newblock \showarticletitle{Preventing Fairness Gerrymandering: Auditing and
  Learning for Subgroup Fairness}. In \bibinfo{booktitle}{\emph{Proceedings of
  the 35th International Conference on Machine Learning, {ICML} 2018}},
  Vol.~\bibinfo{volume}{80}. \bibinfo{publisher}{{PMLR}},
  \bibinfo{pages}{2569--2577}.
\newblock
\urldef\tempurl%
\url{http://proceedings.mlr.press/v80/kearns18a.html}
\showURL{%
\tempurl}


\bibitem[Luo et~al\mbox{.}(2024)]%
        {LuoHYHH0024}
\bibfield{author}{\bibinfo{person}{Renqiang Luo}, \bibinfo{person}{Huafei
  Huang}, \bibinfo{person}{Shuo Yu}, \bibinfo{person}{Zhuoyang Han},
  \bibinfo{person}{Estrid He}, \bibinfo{person}{Xiuzhen Zhang}, {and}
  \bibinfo{person}{Feng Xia}.} \bibinfo{year}{2024}\natexlab{}.
\newblock \showarticletitle{{FUGNN:} Harmonizing Fairness and Utility in Graph
  Neural Networks}. In \bibinfo{booktitle}{\emph{Proceedings of the 30th {ACM}
  {SIGKDD} Conference on Knowledge Discovery and Data Mining, {KDD} 2024}}.
  \bibinfo{pages}{2072--2081}.
\newblock
\href{https://doi.org/10.1145/3637528.3671834}{doi:\nolinkurl{10.1145/3637528.3671834}}


\bibitem[Moro and Cortez(2014)]%
        {bank_marketing_222}
\bibfield{author}{\bibinfo{person}{Rita~P. Moro, S.} {and} \bibinfo{person}{P.
  Cortez}.} \bibinfo{year}{2014}\natexlab{}.
\newblock \bibinfo{title}{{Bank Marketing}}.
\newblock \bibinfo{howpublished}{UCI Machine Learning Repository}.
\newblock
\newblock
\shownote{{DOI}: https://doi.org/10.24432/C5K306}.


\bibitem[Oldfield et~al\mbox{.}(2025)]%
        {OldfieldKXM25}
\bibfield{author}{\bibinfo{person}{Brodie Oldfield}, \bibinfo{person}{Sevvandi
  Kandanaarachchi}, \bibinfo{person}{Ziqi Xu}, {and}
  \bibinfo{person}{Mario~Andr{\'{e}}s Mu{\~{n}}oz}.}
  \bibinfo{year}{2025}\natexlab{}.
\newblock \showarticletitle{An Item Response Theory-based {R} module for
  Algorithm Portfolio Analysis}.
\newblock \bibinfo{journal}{\emph{SoftwareX}}  \bibinfo{volume}{31}
  (\bibinfo{year}{2025}), \bibinfo{pages}{102239}.
\newblock
\href{https://doi.org/10.1016/J.SOFTX.2025.102239}{doi:\nolinkurl{10.1016/J.SOFTX.2025.102239}}


\bibitem[Pleiss et~al\mbox{.}(2017)]%
        {PleissRWKW17}
\bibfield{author}{\bibinfo{person}{Geoff Pleiss}, \bibinfo{person}{Manish
  Raghavan}, \bibinfo{person}{Felix Wu}, \bibinfo{person}{Jon~M. Kleinberg},
  {and} \bibinfo{person}{Kilian~Q. Weinberger}.}
  \bibinfo{year}{2017}\natexlab{}.
\newblock \showarticletitle{On Fairness and Calibration}. In
  \bibinfo{booktitle}{\emph{Advances in Neural Information Processing Systems
  30: Annual Conference on Neural Information Processing Systems 2017}}.
  \bibinfo{pages}{5680--5689}.
\newblock
\urldef\tempurl%
\url{https://proceedings.neurips.cc/paper/2017/hash/b8b9c74ac526fffbeb2d39ab038d1cd7-Abstract.html}
\showURL{%
\tempurl}


\bibitem[ProPublica(2016)]%
        {propublica_compas}
\bibfield{author}{\bibinfo{person}{ProPublica}.}
  \bibinfo{year}{2016}\natexlab{}.
\newblock \bibinfo{title}{ProPublica COMPAS Recidivism Dataset}.
\newblock
\urldef\tempurl%
\url{https://github.com/propublica/compas-analysis}
\showURL{%
\tempurl}


\bibitem[Speicher et~al\mbox{.}(2018)]%
        {speicher2018unified}
\bibfield{author}{\bibinfo{person}{Till Speicher}, \bibinfo{person}{Hoda
  Heidari}, \bibinfo{person}{Nina Grgic{-}Hlaca}, \bibinfo{person}{Krishna~P.
  Gummadi}, \bibinfo{person}{Adish Singla}, \bibinfo{person}{Adrian Weller},
  {and} \bibinfo{person}{Muhammad~Bilal Zafar}.}
  \bibinfo{year}{2018}\natexlab{}.
\newblock \showarticletitle{A Unified Approach to Quantifying Algorithmic
  Unfairness: Measuring Individual {\&}Group Unfairness via Inequality
  Indices}. In \bibinfo{booktitle}{\emph{Proceedings of the 24th {ACM} {SIGKDD}
  International Conference on Knowledge Discovery {\&} Data Mining, {KDD}
  2018}}. \bibinfo{pages}{2239--2248}.
\newblock
\href{https://doi.org/10.1145/3219819.3220046}{doi:\nolinkurl{10.1145/3219819.3220046}}


\bibitem[Weerts et~al\mbox{.}(2023)]%
        {bird2020fairlearn}
\bibfield{author}{\bibinfo{person}{Hilde J.~P. Weerts},
  \bibinfo{person}{Miroslav Dud{\'{\i}}k}, \bibinfo{person}{Richard Edgar},
  \bibinfo{person}{Adrin Jalali}, \bibinfo{person}{Roman Lutz}, {and}
  \bibinfo{person}{Michael Madaio}.} \bibinfo{year}{2023}\natexlab{}.
\newblock \showarticletitle{Fairlearn: Assessing and Improving Fairness of {AI}
  Systems}.
\newblock \bibinfo{journal}{\emph{Journal of Machine Learning Research}}
  \bibinfo{volume}{24} (\bibinfo{year}{2023}), \bibinfo{pages}{257:1--257:8}.
\newblock
\urldef\tempurl%
\url{https://jmlr.org/papers/v24/23-0389.html}
\showURL{%
\tempurl}


\bibitem[Xu et~al\mbox{.}(2025a)]%
        {XuKON25}
\bibfield{author}{\bibinfo{person}{Ziqi Xu}, \bibinfo{person}{Sevvandi
  Kandanaarachchi}, \bibinfo{person}{Cheng~Soon Ong}, {and}
  \bibinfo{person}{Eirini Ntoutsi}.} \bibinfo{year}{2025}\natexlab{a}.
\newblock \showarticletitle{Fairness Evaluation with Item Response Theory}. In
  \bibinfo{booktitle}{\emph{Proceedings of the {ACM} on Web Conference 2025,
  {WWW} 2025}}. \bibinfo{pages}{2276--2288}.
\newblock
\href{https://doi.org/10.1145/3696410.3714883}{doi:\nolinkurl{10.1145/3696410.3714883}}


\bibitem[Xu et~al\mbox{.}(2023)]%
        {XuLCLLW23}
\bibfield{author}{\bibinfo{person}{Ziqi Xu}, \bibinfo{person}{Jixue Liu},
  \bibinfo{person}{Debo Cheng}, \bibinfo{person}{Jiuyong Li},
  \bibinfo{person}{Lin Liu}, {and} \bibinfo{person}{Ke Wang}.}
  \bibinfo{year}{2023}\natexlab{}.
\newblock \showarticletitle{Disentangled Representation with Causal Constraints
  for Counterfactual Fairness}. In \bibinfo{booktitle}{\emph{Advances in
  Knowledge Discovery and Data Mining - 27th Pacific-Asia Conference on
  Knowledge Discovery and Data Mining, {PAKDD} 2023}},
  Vol.~\bibinfo{volume}{13935}. \bibinfo{pages}{471--482}.
\newblock
\href{https://doi.org/10.1007/978-3-031-33374-3\_37}{doi:\nolinkurl{10.1007/978-3-031-33374-3\_37}}


\bibitem[Xu et~al\mbox{.}(2025b)]%
        {XuMRC0X25}
\bibfield{author}{\bibinfo{person}{Ziqi Xu}, \bibinfo{person}{Chenglong Ma},
  \bibinfo{person}{Yongli Ren}, \bibinfo{person}{Jeffrey Chan},
  \bibinfo{person}{Wei Shao}, {and} \bibinfo{person}{Feng Xia}.}
  \bibinfo{year}{2025}\natexlab{b}.
\newblock \showarticletitle{Towards Better Evaluation of Recommendation
  Algorithms with Bi-directional Item Response Theory}. In
  \bibinfo{booktitle}{\emph{Companion Proceedings of the {ACM} on Web
  Conference 2025, {WWW} 2025}}. \bibinfo{pages}{1455--1459}.
\newblock
\href{https://doi.org/10.1145/3701716.3715540}{doi:\nolinkurl{10.1145/3701716.3715540}}


\bibitem[Xu et~al\mbox{.}(2022)]%
        {XuXLCLLW22}
\bibfield{author}{\bibinfo{person}{Zhenlong Xu}, \bibinfo{person}{Ziqi Xu},
  \bibinfo{person}{Jixue Liu}, \bibinfo{person}{Debo Cheng},
  \bibinfo{person}{Jiuyong Li}, \bibinfo{person}{Lin Liu}, {and}
  \bibinfo{person}{Ke Wang}.} \bibinfo{year}{2022}\natexlab{}.
\newblock \showarticletitle{Assessing Classifier Fairness with Collider Bias}.
  In \bibinfo{booktitle}{\emph{Advances in Knowledge Discovery and Data Mining
  - 26th Pacific-Asia Conference, {PAKDD} 2022}}, Vol.~\bibinfo{volume}{13281}.
  \bibinfo{pages}{262--276}.
\newblock
\href{https://doi.org/10.1007/978-3-031-05936-0\_21}{doi:\nolinkurl{10.1007/978-3-031-05936-0\_21}}


\bibitem[Zemel et~al\mbox{.}(2013)]%
        {fairrepresentations}
\bibfield{author}{\bibinfo{person}{Richard~S. Zemel}, \bibinfo{person}{Yu Wu},
  \bibinfo{person}{Kevin Swersky}, \bibinfo{person}{Toniann Pitassi}, {and}
  \bibinfo{person}{Cynthia Dwork}.} \bibinfo{year}{2013}\natexlab{}.
\newblock \showarticletitle{Learning Fair Representations}. In
  \bibinfo{booktitle}{\emph{Proceedings of the 30th International Conference on
  Machine Learning, {ICML} 2013}}, Vol.~\bibinfo{volume}{28}.
  \bibinfo{pages}{325--333}.
\newblock
\urldef\tempurl%
\url{http://proceedings.mlr.press/v28/zemel13.html}
\showURL{%
\tempurl}


\bibitem[Zhao et~al\mbox{.}(2025)]%
        {zhao2025fairdrl}
\bibfield{author}{\bibinfo{person}{Sichen Zhao}, \bibinfo{person}{Wei Shao},
  \bibinfo{person}{Jeffrey Chan}, \bibinfo{person}{Ziqi Xu}, {and}
  \bibinfo{person}{Flora Salim}.} \bibinfo{year}{2025}\natexlab{}.
\newblock \showarticletitle{FairDRL-ST: Disentangled Representation Learning
  for Fair Spatio-Temporal Mobility Prediction}.
\newblock \bibinfo{journal}{\emph{arXiv preprint arXiv:2508.07518}}
  (\bibinfo{year}{2025}).
\newblock


\end{thebibliography}

\end{document}